\documentclass[runningheads]{llncs}
\usepackage{xcolor}
\usepackage{graphicx}
\usepackage{booktabs}
\usepackage{adjustbox}
\usepackage{amsmath}
\usepackage{amssymb}
\usepackage{caption}
\usepackage{subcaption}
\usepackage[misc]{ifsym}
\usepackage{colortbl}
\begin{document}
\title{Efficient In-Context Medical Segmentation with Meta-driven Visual Prompt Selection}

%
\titlerunning{Abbreviated paper title}
%

\author{
    Chenwei Wu\inst{1} \and
    David Restrepo\inst{2} \and
    Zitao Shuai\inst{1} \and \\
    Zhongming Liu\inst{1} \and
     Liyue Shen\inst{1} \Letter
}
\institute{
    University of Michigan, Ann Arbor, United States\\
     \and
    Massachusetts Institute of Technology, Cambridge, United States\\
    \Letter \email{liyues@umich.edu}
}

\authorrunning{C. Wu et al.}
%
\titlerunning{In-Context Medical Segmentation with Meta-driven Prompt Selection}

%
\maketitle              
%

\begin{abstract}
 In-context learning (ICL) with Large Vision Models (LVMs) presents a promising avenue in medical image segmentation by reducing the reliance on extensive labeling. However, the ICL performance of LVMs highly depends on the choices of visual prompts and suffers from domain shifts. While existing works leveraging LVMs for medical tasks
 have focused mainly on \textit{model-centric} approaches like fine-tuning, we study an orthogonal \textit{data-centric} perspective on how to select good visual prompts to facilitate generalization to medical domain. 
 In this work, we propose a label-efficient in-context medical segmentation method by introducing a novel \emph{\textbf{M}eta-driven \textbf{V}isual \textbf{P}rompt \textbf{S}election} mechanism (\textbf{MVPS}),
 where a prompt retriever obtained from a meta-learning framework actively selects the optimal images as prompts to promote model performance and generalizability.
 Evaluated on 8 datasets and 4 tasks across 3 medical imaging modalities, our proposed approach demonstrates consistent gains over existing methods under different scenarios, improving both computational and label efficiency. Finally, we show that MVPS is a flexible, finetuning-free module that could be easily plugged into different backbones and combined with other model-centric approaches.

\end{abstract}
\section{Introduction}
Large vision models (LVMs), such as SegGPT~\cite{wang2023seggpt}, Painter~\cite{wang2023images}, and variants of SAM~\cite{kirillov2023sam}, can learn to perform segmentation on unseen tasks from a few visual prompts without updating model parameters. This in-context learning (ICL) ability enables label-efficient medical image segmentation~\cite{wu2023medical,deng2023segment} without relying on expensive expert annotations and thus ease the deployment of medical AI.

\begin{figure}[t]
    \centering
    \includegraphics[width=0.9\linewidth]{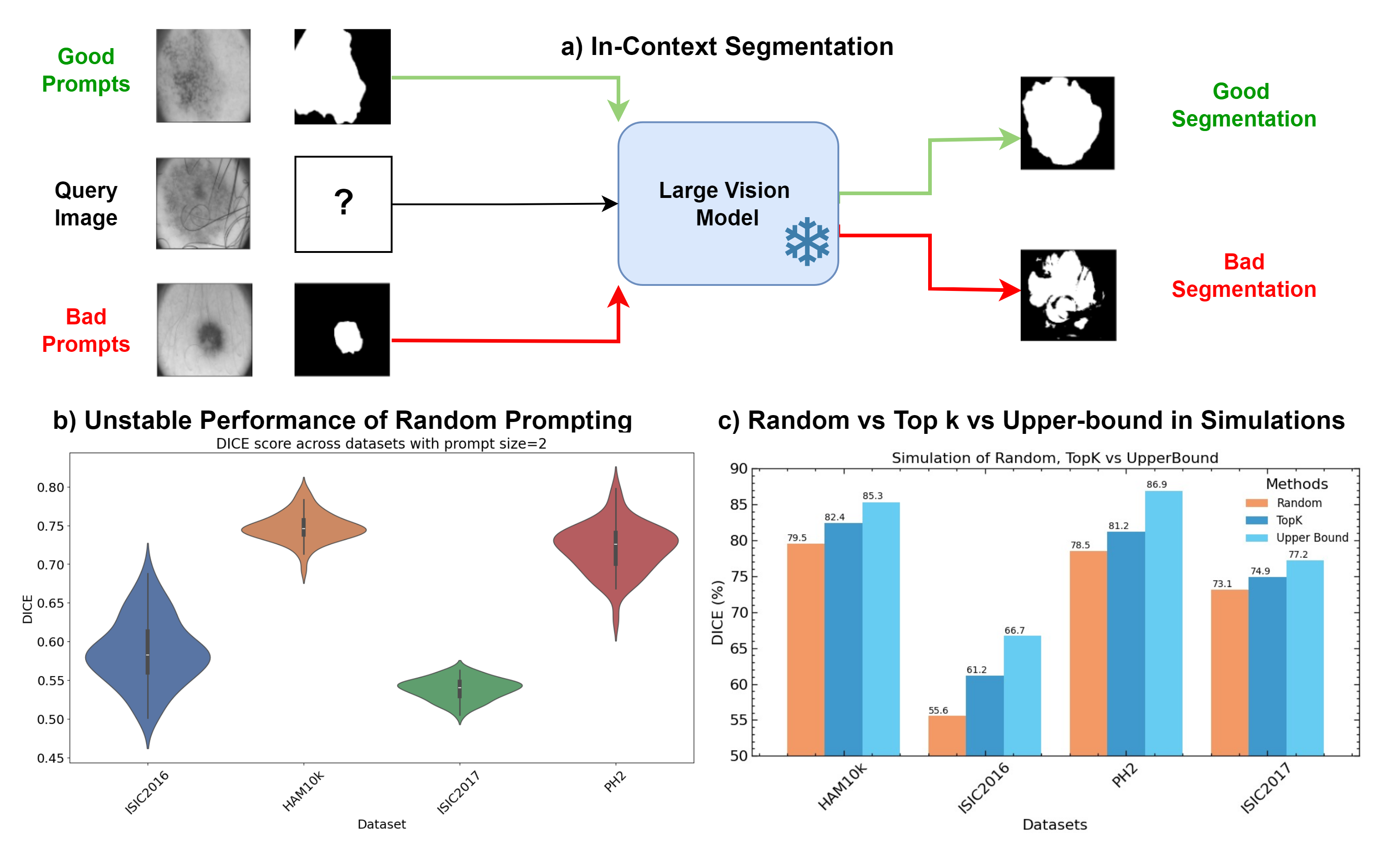}
    \caption{\textbf{a) In-context segmentation.} Large Vision Models are capable of taking in visual prompts of image-mask pairs and outputting the segmentation mask prediction for the query image. 
    \textbf{b) Instability of ICL with random prompting.} ICL has shown unstable performance using random prompts. By conducting experiments of in-context skin lesion segmentation using SegGPT~\cite{wang2023seggpt} on 4 different dermatology datasets~\cite{tschandl2018ham10000,isicdatasetcodella2018skin,mendoncca2015ph2}, with prompt size of 2,  the results show a large variance. (Mean DICE scores: $37.36\%$, $14.84\%$,$11.46\%$, and $31.80\%$).
    \textbf{c) Better prompts lead to a significant improvement in ICL.} In this simulation study, we iterate through all prompt selection options (with prompt size of 2 in this example) given a prompt pool of 100 images and test ICL performance. There is plenty of room for improvement over the current prompt selection methods like TopK \cite{zhang2023whatmakesgoodexamples} approach.}
    \label{fig:1}
\end{figure}

However, the effectiveness of ICL depends on the quality and number of visual prompts~\cite{r11,zhang2023whatmakesgoodexamples}. Transferring LVMs from natural images to medical images faces further challenges due to domain shifts~\cite{deng2023segment} and the large variability across patient populations, demographics, imaging protocols, etc.
For example, Figure~\ref{fig:1}(b) demonstrates this issue in dermatology image segmentation. Using an LVM with two randomly selected visual prompts, the performance varies by 20.11\% on average across four different dermatology image datasets~\cite{tschandl2018ham10000,isicdatasetcodella2018skin,mendoncca2015ph2,gutman2016skin}. ICL also demonstrates high sensitivity to the number of prompts in challenging segmentation tasks such as vessel segmentation.

While the critical role of prompt selection has gained significant attention in the Large Language Model(LLM) field~\cite{zhang2022activenlp}, LVM adaptation works have been focusing primarily on the model-centric approaches such as efficient model fine-tuning~\cite{hu2021lora,zhu2023melo}. Some works~\cite{rubin2021learning,lu2021fantastically} have explored the impacts of prompt quality on LLM adaptation to new data and tasks;~\cite{mavromatis2023examples,zhang2022activenlp} focuses on formulating the selective prompt labeling as a data-centric method to determine which prompt samples are valuable for labeling, similar to active learning. On the imaging side,~\cite{ma2024segment} adapts LVMs to medical images by fine-tuning the model on extensive medical images, but the complexity of training large-scale models and the need for massive and expensive labeled data highlight the significant resources required for such comprehensive models;~\cite{wu2023medical} utilizes an adaptor to fine-tune LVMs efficiently to medical datasets but remains model-centric.~\cite{zhang2023whatmakesgoodexamples} is the first work to discuss visual prompt selection on natural images and proposes an unsupervised TopK approach and a supervised method SupPR for prompt selection. SupPR requires training of a large visual embedding extractor and demonstrating slight improvements over TopK approach. Other works have also explored how to improve the prompt quality given a fixed set of prompts:~\cite{r11} proposes prompt fusion, using an ensemble of different prompt layouts to activate knowledge at different positions in the LVMs;~\cite{r12} devises a prompt enhancer trained on large-scale labeled data to add perturbations to the in-context prompt pairs. However, these works still rely on TopK prompt selection, which are constrained by the distance measurements and cannot always find the visual prompt optimal across domains. 
Overall, there hasn't been a generic solution to optimizing visual prompt selection to adapt LVMs efficiently for in-context medical segmentation.


Herein, we introduce MVPS, a novel Meta-learning-driven Visual Prompt Selection framework. The key contributions of our research lie in three-fold: 
\begin{enumerate}
    \item We propose a label-efficient in-context medical segmentation method enabled by introducing a novel Meta-driven Visual Prompt Selection mechanism (MVPS), which enables cross-domain adaptation of LVMs to medical imaging. 
    \item Specifically, we propose a meta-driven active visual prompt retrieval approach, by constructing a meta-learning scheme to teach a transformer-based prompt retriever which images are worthy of being selected as visual prompts to boost model performance. The prompt retriever is optimized through the probability distribution estimation and reward policy gradients.
    \item MVPS provides a data-centric and finetuning-free enhancement to in-context medical segmentation, with both data and label efficiencies largely benefitting medical applications. 
    MVPS consistently outperforms baselines across 8 datasets and 4 tasks across 3 medical imaging modalities. We also show that MVPS is flexible enough to plug into different LVM backbones, and to be combined with model-centric approaches.

\end{enumerate}

\section{Methodology}

\begin{figure}
    \centering
    \includegraphics[width=0.9\linewidth]{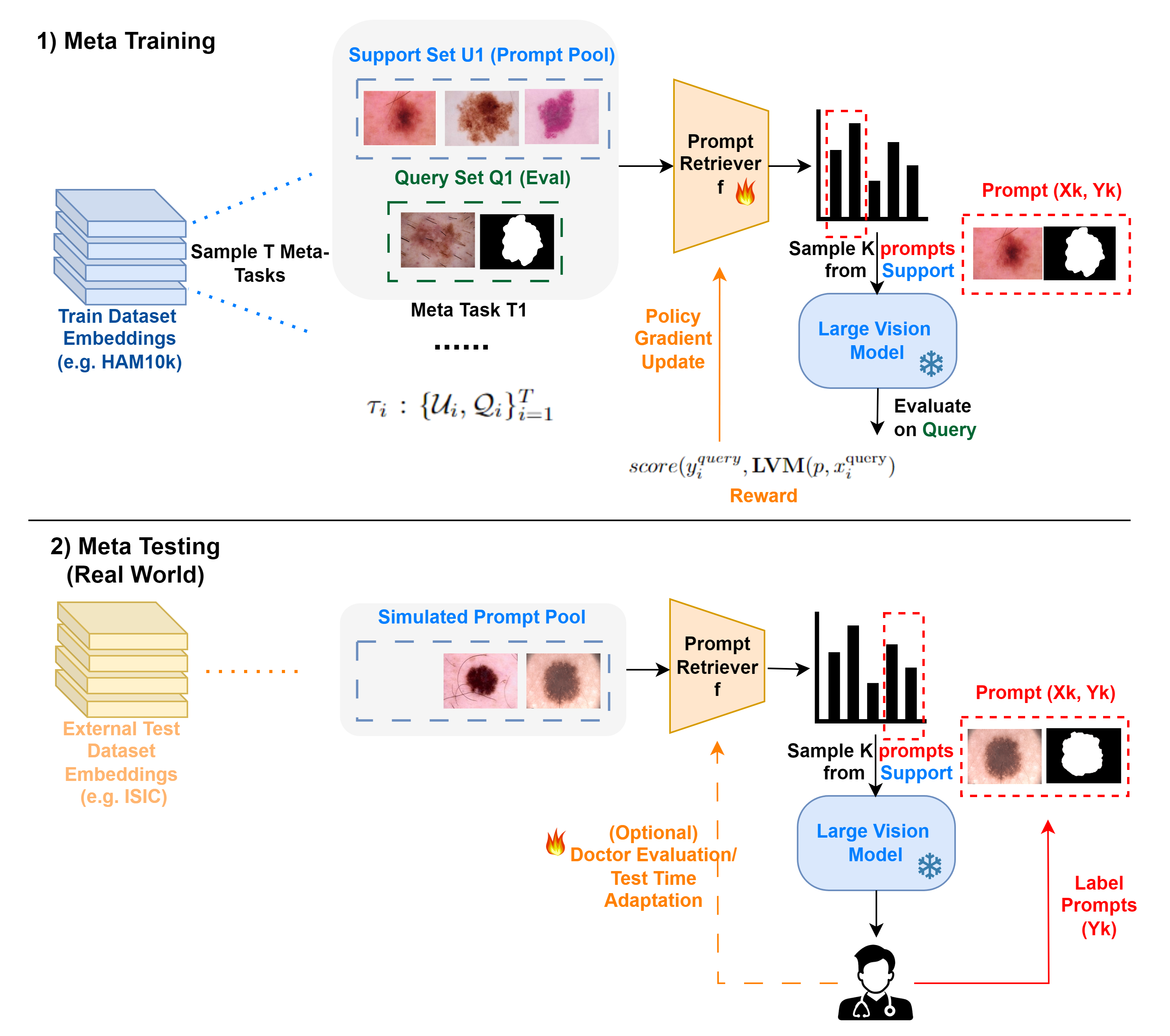}
  \caption{Meta-training and meta-testing stages of the proposed MVPS framework (use dermatology dataset as an example). Note that prompt retriever is trainable while large vision model is kept frozen.}
    \label{fig:fig3}
\end{figure}

In the following section, we will introduce the technical details of the proposed MVPS framework as shown in Fig \ref{fig:fig3}. Taking Dermatology as an example, given a training set (e.g., HAM10k) and a testing set (e.g., ISIC), the MVPS framework constructs tasks for meta-training and meta-testing. Each task consists of a \textbf{Support Set} of 1000 unlabeled images and a \textbf{Query Set} of 100 labeled image-mask pairs. The goal of meta-training is to learn a prompt retriever \( f(\cdot) \) that can select the best prompts from the support set to enhance segmentation performance on the query set. The retriever is optimized by comparing the output segmentation masks of the Large Vision Model (LVM) against the ground truth masks in the query set, and then using this feedback to improve prompt selection. For meta-testing, the retriever selects prompts from a simulated prompt pool of 1000 images in the test dataset. The LVM uses these prompts to perform segmentation on new query images. 
During meta-testing, we evaluate and report the LVM's segmentation performance on different target (meta-test) datasets in order to mimic the real-world scenario when the LVMs are utilized in a different hospital with a new patient cohort. 

\subsubsection{Meta Tasks Construction.} 
As shown in Fig.~\ref{fig:fig3}, in our meta-learning setting, each task is an in-context medical segmentation problem, denoted as $\tau_i: \{ \mathcal{U}_i, \mathcal{Q}_i\}_{i=1}^{T}$.
Suppose there are a total of $T$ tasks.
Each task consists of an unlabeled image set $\mathcal{U} = \{ x_j \}_{j=1}^{1000}$, which we refer to as the support set or prompt pool, and a labeled image set $\mathcal{Q} = \{ (x_k, y_k) \}_{k=1}^{100}$
which we refer to as query set consisting of a set of image $x$ and segmentation label $y$. 
Both $\mathcal{Q}$ and $\mathcal{U}$ are sampled from the training datasets and there is no overlap between the images in $\mathcal{Q}$ and $\mathcal{U}$ within a single task. 
In this way, we define each meta prompt learning task $\tau_i$ as:
$
\tau_i=\left(\mathcal{U}_{\tau_i}^{\text {support }}, \mathcal{Q}_{\tau_i}^{\text {query }}\right),
$
where $\mathcal{U}_{\tau_i}^{\text {support }}$ indicates the support set as the candidate pool for prompt selection, and $\mathcal{Q}_{\tau_i}^{\text {query}}$ indicates the query set to evaluate the in-context learning performance for the current task $\tau_i$. 
Following the setup of active learning, we hold out specific disease categories if disease category labels are available. We only sample the task $\tau_i$ from the non-heldout labels for the meta-training tasks. For meta-validation, we sample from a mix of non-heldout labels and held-out labels. 
This ensures that meta-training tasks and meta-validation tasks are not sampled from too similar domains to prevent the model from simply memorizing training samples.

\subsubsection{Prompt Retriever.}
Following previous active learning research~\cite{rauch2023activeglae}, we design a trainable prompt retriever using the transformer architecture. The retriever takes in a pair of extracted embeddings for both unlabeled and labeled images$\left(\mathcal{U}_{\tau_i}^{\text {support }}, \mathcal{Q}_{\tau_i}^{\text {query }}\right)$ and outputs a softmax-normalized distribution which indicates the probability of all the samples in the prompt pool to be selected as a good prompt over the prompt pool samples. 
In order to distinguish the support and query representations \cite{geng2022multimodal}, we add two learnable vectors that represent prompt pool and query sets, respectively, to their corresponding embeddings. 
Based upon this predicted distribution, we select $k$ samples with the highest probabilities, along with their labels, as the visual prompts. These image-mask pairs will serve as inputs to the LVMs for completing in-context segmentation tasks to predict the segmentation masks for query images.

\subsubsection{Task Augmentation for Better Generalization.}
To perform more robust meta-training and address the data scarcity problem in domains with small training sets, we propose a new task augmentation by mixing up images and segmentation masks.
Specifically, we extend the idea of mix-ups to task augmentations to further densify the task distribution. 
Randomly selecting an anchor task $\tau_i$, and another task $\tau_j$, we form a new task and support set by interpolating the $\left(x_{\tau_i}^{\text {support }}, x_{\tau_j}^{\text {support }}\right)$ and corresponding masks  $(y_{\tau_i}^{\text {support }}, y_{\tau_j}^{\text {support }})$ from $\tau_i$ and $\tau_j$, using a mixing ratio $\lambda \in[0,1]$, so that $\tilde{x}_{\tau_i} = (1-\lambda) x_{\tau_i} + \lambda x_{\tau_j}, \quad \tilde{y}_{\tau_i} = (1-\lambda) y_{\tau_i} + \lambda y_{\tau_j}$.
\subsubsection{Optimization and Reward Shaping.}
The reward \( \mathcal{R} \) is determined by a scoring function that evaluates the efficacy of the Large Vision Model (LVM) on the query set $\mathcal{Q}_{\tau_i}$, when given a selected prompt \( p \) from $\mathcal{U}$, formally denoted as
\begin{equation}
\footnotesize
\mathcal{R}(p,\mathcal{Q}_{\tau_i}^{\text{query}}) = score(y_i^{query},\mathbf{LVM}(p,x_i^{\text{query}})), x_i,y_i \in \mathcal{Q}_{\tau_i}^{\text{query}}
\end{equation}

\noindent Here, we employ the DICE coefficient on the query dataset to calculate this reward.  
The aim is to train the prompt retriever to act as a policy model to discern and generalize the characteristics of good contextual prompts and enable the selection of informative unlabeled examples in scenarios where the retriever may not have exposure to any labeled instances. We utilize the simple but effective expected reinforcement performance as our objective~\cite{williams1992simple,zhang2022activenlp,smit2021medselect} and optimize by back-propagation of policy-gradients 
\begin{equation}
\nabla_\theta \mathcal{L}=\mathbb{E}_{ \sim \mathbb{P}_\theta(p \mid \mathcal{U}_{\tau_i}^{\text {support}})}[\left(\mathcal{R}\left(p, \mathcal{Q}_{\tau_i}^{\text {query }}\right)) \nabla_\theta \log \left(\mathbb{P}_\theta(p \mid \mathcal{U}_{\tau_i}^{\text {support}})\right)\right]
\end{equation} \label{eq:loss}
\noindent Here we approximate the expected value by a single Monte-Carlo sample. Previous works in reinforcement learning~\cite{zhang2022activenlp,chen2023self} have found that sparse reward learning is unstable and difficult. Reshaping the reward function into an intuitive additional gain of getting the retrieved sample will help stabilize the training while preserving
the invariance of optimal policies. Therefore, we utilize the reshaped reward function based on the marginal utility our $p$ gains over the prompt $p^{\prime}$ random baseline retrieves: $\mathcal{R}(p)=\mathcal{R}(p,\mathcal{Q}_{\tau_i}^{\text{query}})-\mathcal{R}(p^{\prime},\mathcal{Q}_{\tau_i}^{\text{query}})$.

\section{Results}
To evaluate the performance of the proposed MVPS method, we conduct experiments on three medical image modalities (dermatology, ophthalmology, and radiology) and four tasks (skin lesion segmentation, lung lesion segmentation, optical disc segmentation, and vessel segmentation) over 11 datasets.

\noindent \textbf{Datasets.}
For \textit{dermatology image segmentation} task:
we use the HAM10k~\cite{tschandl2018ham10000} dataset for meta-training and validation; we test on PH2~\cite{mendoncca2015ph2} and ISIC2016~\cite{gutman2016skin} datasets to study the relationship between prompt retrieval effectiveness and \textbf{distribution shifts} between source and target domains, including task/label shifts, population shifts, and technology shifts.
For \textit{ophthalmology image segmentation task}: we design two subtasks to explore prompt retrieval behavior under \textbf{contrasting task difficulties}. For optical disc segmentation, an easier task, we train on the Refuge~\cite{orlando2020refuge} dataset and test on a combined set from PAPILA~\cite{kovalyk2022papila}+IDRID~\cite{porwal2018indian}. For the more challenging vessel segmentation task, we use the FIVES~\cite{jin2022fives} dataset as the source, a combined DRIVE~\cite{staal2004ridgedrive}, CHASEDB~\cite{fraz2012ensemblechasedb}, and STARE~\cite{hoover2000locating} datasets as the test sets. 
For \textit{Chest X-ray COVID-19 identification task}: we use the CovidQUEX dataset as train set. Collected by the same group, the COVID-19 Radiography dataset~\cite{chowdhury2020can} is used for meta-testing, examining a scenario where the source and target share the \textbf{same task} and minimal shifts.

\noindent\textbf{Experiments and Comparisons.}
For prompt retriever, we use a transformer with an embedding size of 256, a max sequence length of 1101 (including 1000 support size, 100 query size, and 1 separator token), 8 heads, 8 encoder blocks, 4 decoder blocks, and a feedforward dimension of 2048. We sample 150000 meta-tasks for each modality and use a mix-up ratio of 0.1 for augmentation. We use an Adam optimizer with a learning rate of 1e-4 and batch size of 64 for 10 epochs. We use SegGPT \cite{wang2023seggpt} as the LVM backbone and a LAION-2B ViT model for embedding extraction. Experiments are done on 1 AWS-g5.12xlarge instance. 

\noindent We compare against random selection, TopK/Unsupervised approachs and the SupPR~\cite{zhang2023whatmakesgoodexamples} as data-centric prompt retrieval approaches. We also compare with model-centric low-rank adaptation (LoRA) methods~\cite{hu2021lora,zhu2023melo} for LVM adaptation to medical domain, where we finetune 14\% of total params on the train set with full labels. 
To avoid randomness, we run experiments of 30 runs and report average scores as results.
Sampling at a step size of $k=1$, we test the regular MVPS frozen during test time and MVPS with test time adaptation (TTA) receiving policy gradient updates during test time, which assumes initial access to a dynamically expanding tiny labeled evaluation set as more prompt images get labeled. We also show that MVPS is a flexible module that can be combined with model-centric approaches like LoRA. Finally, we approximate a performance upper bound using supervised SOTAs~\cite{wu2023medical}.

\begin{figure}
    \centering
    \includegraphics[width=0.9\linewidth, height = 6cm]{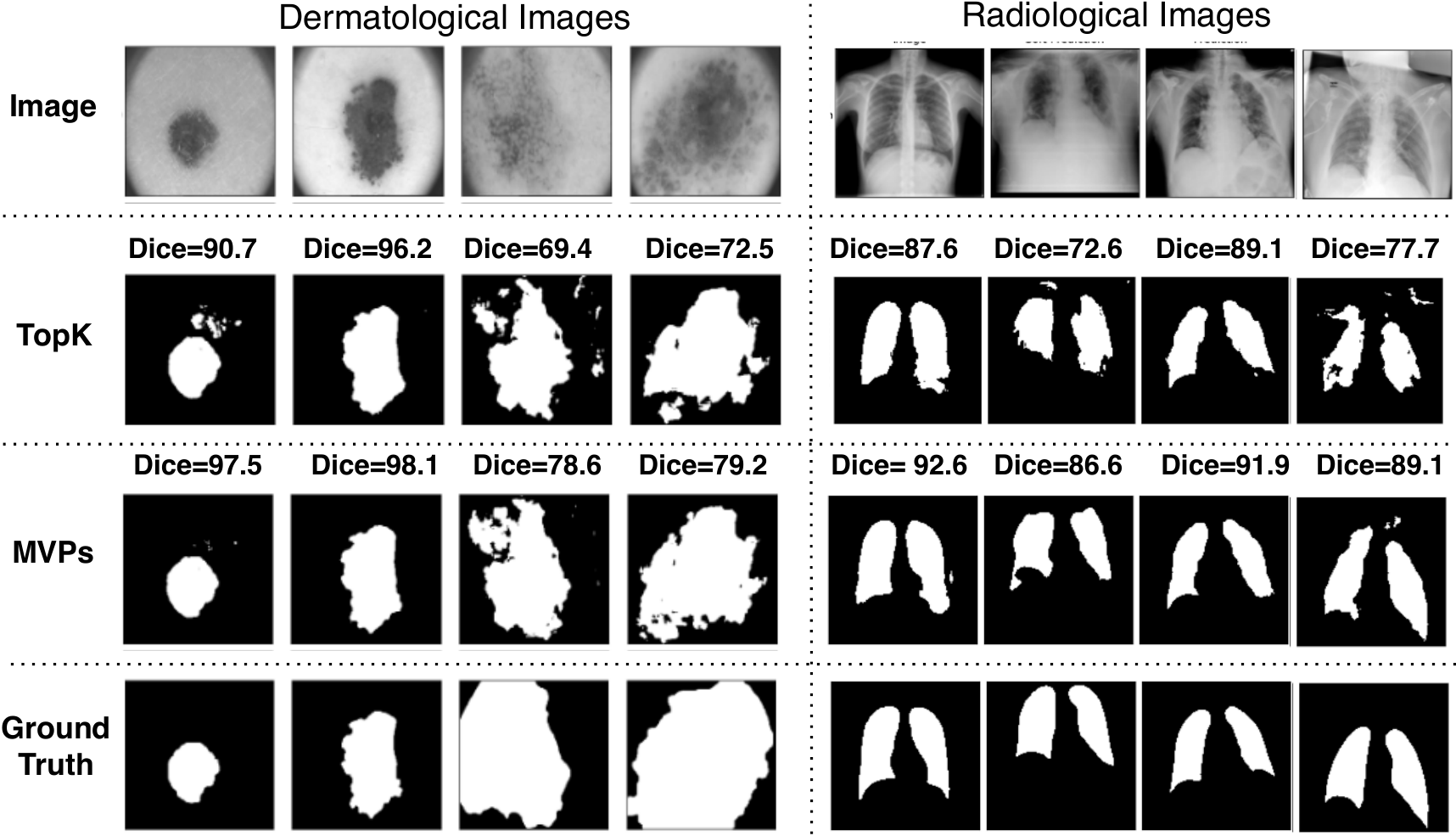}
    \caption{Segmentation Results from MVPS vs TopK Prompting.}
    \label{fig:qualitative}
\end{figure}

\noindent \textbf{Findings.} Tested at prompt length $k=2, 4, 8, 16, 32$ (for retinal images, we stop at 16 due to small test size), MVPS offers consistent ICL enhancements under various source-target settings, including the same task, different task difficulties, and different distribution shifts. Qualitatively, we found that better prompts found by the retriever help the model better recognize noisy patterns adjacent to lesions and not just segment based on object contrasts, as shown in Fig \ref{fig:qualitative}. Quantitatively, MVPS and MVPS+TTA average a 4.08\% and 4.66\% gain over TopK and a 7.40\% and 7.97\% gain over random selection. For the \textbf{same task} and minimal distribution shifts setting in X-rays, we observe a 3.6\% performance gain of the frozen LVM over TopK but diminishing returns as the number of k increases. When we combine MVPS with LoRA, it reaches a 95.1\% comparable to supervised SOTA at $k=32$. 
\begin{table*} 
\centering 
\label{tab:displaymain}

\scalebox{1}{
\begin{tabular}{@{} ll | l | lll |l|ll }
\toprule
\textbf{Method} & \textbf{Type} &\textbf{K} &\textbf{PH2} &\textbf{ISIC} &\textbf{Xray} &\textbf{K} &\textbf{Optical} &\textbf{Vessel} \\
\midrule
\midrule
 Random&Data &4 & 79.8 & 63.1 & 77.8 & 4& 46.1 & 41 \\
 & & 8 & 80.1 & 67.9 & 82.1 & 8& 64.4 & 59 \\
 & & 32 & 81.5 & 70.5 & 87.2& 16& 73 & 62.3\\

\midrule
 TopK~\cite{r11,r12,zhang2023whatmakesgoodexamples}&Data &4 & 84.0 & 67.2 & 79.0 & 4& 55 & 44.0 \\
 & & 8 & 84.2 & 72.3 & 84.3 & 8& 65.5 & 61.0 \\
 & & 32 & 85.8 & 76.5 & 88.3 & 16 & 73.5 & 64.5 \\

\midrule

SupPR~\cite{zhang2023whatmakesgoodexamples} & Data &4 & 88.17 & 72.69 & 81.95 & 4 & 59.77 & 47.97 \\
 & & 8 & 88.31 & 76.18 & 85.88 & 8 & 73.43 & 70.65\\
   & & 32 & 89.28 & 77.20 & 88.10 & 16 & 79.22 & 72.53\\
\midrule
 MVPS & Data &4 & \textbf{88.2}$_{\uparrow 4.2}$& \textbf{72.3}$_{\uparrow 5.1}$& \textbf{82.6}$_{\uparrow 3.6}$ & 4 &\textbf{59.9}$_{\uparrow 4.9}$& \textbf{48.6}$_{\uparrow 4.6}$\\
& & 8 & \textbf{88.5}$_{\uparrow 4.3}$ & \textbf{76.2}$_{\uparrow 3.6}$ & \textbf{86.2}$_{\uparrow 1.9}$ & 8 & \textbf{73.6}$_{\uparrow 8.1}$& \textbf{71.0}$_{\uparrow 10.0}$\\
& & 32 & \textbf{89.5}$_{\uparrow 3.7}$ & \textbf{77.9}$_{\uparrow 1.4}$& \textbf{89.1}$_{\uparrow 0.8}$ &16 & \textbf{79.9}$_{\uparrow 6.4}$ &\textbf{72.6}$_{\uparrow 7.1}$  \\

\midrule
\midrule
LoRA~\cite{zhu2023melo,hu2021lora} & Model &4 & 85.2 & 75.6 & 83.8 & 4 &73.4  & 61.2\\
 & & 8 & 86.3 & 78.2 & 88.8 & 8 & 79.3  & 72.5\\
   & & 32 & 91.5 & 82.3 & 93.3 &16 & 86.9 & 74.6\\
\midrule
\midrule
LoRA+MVPS & Combined &4 & \textbf{89.9}$_{\uparrow 4.7}$& \textbf{78.5}$_{\uparrow 2.9}$& \textbf{85.9}$_{\uparrow 2.1}$ &  4 & \textbf{76.5}$_{\uparrow 3.1}$&  \textbf{68.6}$_{\uparrow 7.4}$\\
& & 8 & \textbf{90.3}$_{\uparrow 4.0}$& \textbf{79.4}$_{\uparrow 1.2}$& \textbf{89.8}$_{\uparrow 1.0}$ & 8 & \textbf{81.7}$_{\uparrow 2.4}$ & \textbf{75.3}$_{\uparrow 2.8}$\\
 & & 32 & \textbf{93.3}$_{\uparrow 0.8}$ & \textbf{92.3}$_{\uparrow 2.8}$ & \textbf{95.1}$_{\uparrow 1.8}$ & 16& \textbf{89.7}$_{\uparrow2.8}$ & \textbf{75.7}$_{\uparrow1.1}$\\
\midrule
\midrule
\rowcolor{gray!50}
Supervised~\cite{wu2023medical} & - & - & 96.4 & 94.8 & 98.1 & & 96.6 & 86.8\\

\midrule
\bottomrule
\end{tabular}
}
\caption{DICE Score Over Different Datasets and K. Full results in Appendix. DICE was used as both our scoring function and performance metric, for the convenience of showing the score of supervised methods as an upper bound for direct comparison. We have also tested the mIOU as scoring function and observed similar patterns.}
\end{table*}

\begin{table*} 
\centering 
\label{tab:displaymain}

\scalebox{1}{
\begin{tabular}{@{} ll|l | lll|l|ll }
\toprule
\textbf{Method} & \textbf{Type} &\textbf{K} &\textbf{PH2} &\textbf{ISIC} &\textbf{Xray} &\textbf{K} &\textbf{Optical} &\textbf{Vessel} \\
\midrule
 MVPS & Data &4 & 88.2 & 72.3 & 82.6 & 4 &59.9 & 48.6\\
& & 8 & 88.5 & 76.2 & 86.2 & 8 & 73.6 & 71.0\\
& & 32 & 89.5 & 77.9 & 89.1 &16 & 79.9 &72.6 \\
\midrule
 MVPS+TTA&Data & 4 & 88.2 & 72.8 & 85.4& 4 & 59.9 & 48.6\\
& & 8 & 88.5 & 76.2 & 88.2 & 8 & 73.6 & 71.0\\
& & 32 & 90.9 & 78.3 & 89.5 &16 & 81.1 & 73.2\\
\bottomrule
\end{tabular}
}
\caption{Ablation: Does Test Time Adaptation Help?}
\end{table*}

For \textbf{contrasting task difficulties} in Opthalmology experiments, we found that the harder task (Vessel) has less (4.87\%) but still significant average gain compared to the easier task (Optical Disc) (6.725\%), similar to those findings in NLP prompting~\cite{zhang2022activenlp} and suggesting that prompt insights generalize across tasks. As shown in Table 1, for different \textbf{distribution shifts} in Dermatology, we found that MVPS shows consistent gains despite population shifts (ISIC) and technology shifts (PH2), while stronger label shifts in ISIC dataset may be the cause of less average gain than PH2 dataset. As an \textbf{ablation}, we found that our test time adaptation indeed helps improve performance further upon MVPS, yet we have smaller gains when the prompt pool is small in PH2 or retinal datasets.

\noindent \textbf{Efficiency} 
With frozen LVMs, MVPS only needs to train the prompt retriever with trainable parameters of around $22M$. For a comparison, parameter-efficient fine-tuning methods like LoRA have 52M trainable parameters, while fully fine-tuning LVMs such as medicalSAM ~\cite{wu2023medical} will require 260M parameters. SupPR requires contrastive pretraining of a vision transformer as a feature extractor (~315 million parameters).
It is also a flexible module that isn't exclusive to other model-centric methods like LoRA, with our results showing a consistent gain of 2.93\% improvement when combined. We also tested MVPS on other in-context learning backbones such as Universeg~\cite{butoi2023universeg} on the PH2 dataset and still got a gain of 3.70\% over TopK and our combined approach 4.92\% over LoRA.

\section{Conclusion}
To enhance ICL efficiency when applying LVMs to new medical imaging domains, we propose MVPS, a Meta-driven Active Prompt Selection framework. We meta-train a transformer-based prompt retriever optimized through reshaped reward policy gradients.
MVPS offers a data-centric and finetuning-free enhancement to ICL, leading to efficiencies in training and label acquisition processes.
MVPS's flexibility allows for integration with different LVM backbones and model-centric approaches.

\subsubsection{\ackname} The authors acknowledge support from Michigan Institute for Computational Discovery and Engineering (MICDE) Catalyst Grant, and Michigan Institute for Data Science (MIDAS) PODS Grant.
\subsubsection{Disclosure of Interests}
The authors have no competing interests to declare that are relevant to the content of this article.

\newpage

%
%
%
%

\end{document}